\documentclass{article}
\usepackage{microtype}
\usepackage{graphicx}
\usepackage{booktabs}
\usepackage{hyperref}
\usepackage{xurl}

\usepackage[preprint]{icml2026}
\usepackage{amsmath}
\usepackage{amssymb}
\usepackage{cleveref}
\newcommand{\method}{MoDDiT}
\icmltitlerunning{MoDDiT: Simple Token Routing in Diffusion Transformers}

\begin{document}
\twocolumn[
\icmltitle{MoDDiT: Simple Token Routing in Diffusion Transformers}
\icmlsetsymbol{equal}{*}
\begin{icmlauthorlist}
\icmlauthor{Jarek Liesen}{equal,bold}
\icmlauthor{Matthew T. Jackson}{equal,bold}
\icmlauthor{Benjamin Ellis}{bold}
\icmlauthor{Shimon Whiteson}{bold}
\icmlauthor{Jakob N. Foerster}{bold}
\end{icmlauthorlist}
\icmlaffiliation{bold}{British Open-ended Learning and Discovery}
\icmlcorrespondingauthor{Jarek Liesen}{jarek.liesen@reuben.ox.ac.uk}
\vskip 0.3in
]
\printAffiliationsAndNotice{\icmlEqualContribution}

\begin{abstract}
Real-time video generation, interactive world models, and on-device image synthesis are almost a reality.
Diffusion Transformers (DiT) achieve excellent sample quality, shifting the bottleneck to the throughput of the models themselves.
Natural images are inherently sparse, as large portions of an image are often background, contribute little to the scene content, and are easy to denoise.
DiTs fail to exploit this sparsity, as they process image tokens statically and uniformly.

We propose \method{}, a simple architecture modification that learns to route tokens through both MLP and self-attention operations.
By routing tokens through both blocks independently, \method{} allows for dynamic compute allocation in MLPs that varies across tokens, images, and denoising timesteps.
Using \method{} to train image generation models from scratch significantly reduces computational costs of training and inference, processing as few as 12.5\% of the tokens. 
By increasing sparsity and model depth, \method{} reduces costs and improves generation performance over unrouted baselines.
The learned allocation is interpretable, revealing how computation varies with visual content and noise level. 
We also use \method{} to train video generation models, where allocation correlates with motion, and to distill a dense bidirectional video model into a sparse autoregressive student.
\end{abstract}

\section{Introduction}
Recent advancements in real-time video generation, interactive world models, and on-device image synthesis promise numerous new applications for generative models, and a new frontier of synthetic data.
Diffusion models are at the forefront of this progress but are bottlenecked by their iterative generation procedure, which requires multiple model evaluations to generate samples.
The resulting latency is often reduced by distilling models into smaller students or subsampling diffusion timesteps.
A complementary axis of improvement is given by architecture modifications that reduce the cost of each inference step of the denoising process.

A promising route is exploiting sparsity, inspired by the observation that images and videos are information-sparse modalities where few patches per frame contain a large amount of information.
For example, in many images the background is much simpler than the foreground and is therefore easier to denoise.
This effect is magnified in videos, where patches containing static objects can be simply copied over from the previous frame. 
Ideally, diffusion would focus computation on more relevant patches, reducing overall latency.
However, predominant diffusion architectures such as diffusion transformers~\citep[DiT,][]{peebles2023scalable} fail to exploit this property, instead applying all operations to every token, thereby allocating compute uniformly over images and videos. 

In this work, we introduce \textbf{\method{}}, a simple mechanism for routing tokens through both MLP and self-attention operations (\cref{sec:routing-mechanism}).
While routing through self-attention uses fixed capacity, routing through MLPs enables dynamic compute allocation across tokens, images, and denoising timesteps (\cref{sec:mlp-routing}).
This allows the model to devote more computation to regions that are visually complex and difficult to denoise.

\begin{figure*}[t]
    \centering
    \includegraphics[width=\linewidth]{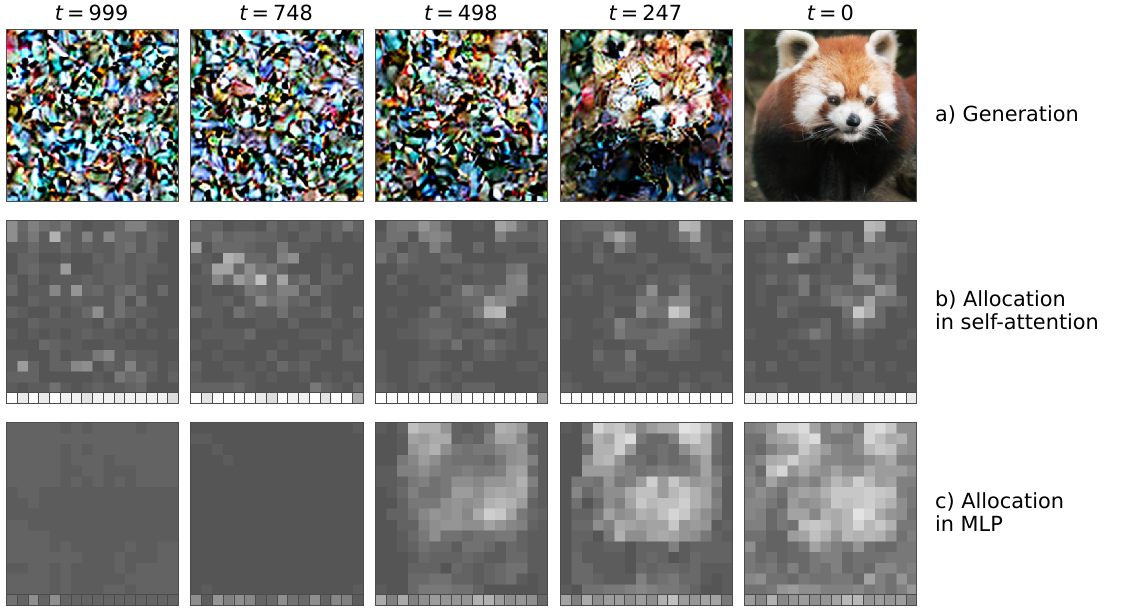}
    \caption{Image generation and learned compute allocation in \method{} for a red panda example. \textbf{(a)} Generated images along the denoising trajectory, from left to right. \textbf{(b)} Self-attention allocation. \textbf{(c)} MLP allocation. Patch brightness indicates the proportion of blocks applied to each token at the given timestep; the strips below the allocation panels show register-token allocation.}
    \label{fig:im-sample}
    
\end{figure*}

Unlike prior methods, \method{} reduces FLOPs during both training and inference, requires no auxiliary routing loss or post-processing, and enables efficient training on parallel GPU hardware without ragged batches.
As illustrated in \cref{fig:im-sample}, the resulting allocation is interpretable, with computation varying across image regions, samples, and denoising timesteps.

We evaluate \method{} across three settings: image generation, video generation, and video distillation.
On ImageNet~\citep{imagenet} (\cref{sec:im-expts}), \method{} improves the quality--compute tradeoff over vanilla DiT across a range of routing capacities and sampling lengths.
For video generation, we evaluate on the 1X World Modeling Challenge~\citep{1x_world_model_raw_data_2025} (\cref{sec:vid-expts}), where routing at 12.5\% capacity improves video-prediction PSNR while requiring 36.92\% fewer counted denoiser FLOPs on the profiled clip than the DiT baseline.
Finally, we integrate sparse feedforward routing into the distillation of pretrained Wan2.1~\citep{wan2025}, converting a dense bidirectional model into a sparse, few-step autoregressive generator (\cref{sec:distillation}).
Compared with a matched dense autoregressive student, the sparse model almost matches the dense baseline with a $5\%$ overall reduction in score on the VBench benchmark \citep{huang2023vbenchcomprehensivebenchmarksuite}, while reducing inference FLOPs by $24\%$.
We further analyze the learned routing behavior (\cref{fig:im-loss-capacity,fig:vid-compute-alloc}), showing how computation varies with denoising stage, image and patch class, and temporal change in video.

In summary, \method{} provides a simple, efficient mechanism for dynamic token routing, exploiting token-level sparsity in diffusion transformers. By increasing model size and sparsity, it can increase performance while reducing training- and inference-time FLOPs.
We compare routing choices and ablations on ImageNet, and give intuitive explanations for allocation across image content and time steps. On 1X, we test action-conditioned video generation and show that learned allocation varies with temporal change. Distilling Wan2.1 tests whether \method{} can be combined with distillation and few-step denoising, where the reduction in inference FLOPs comes with a modest reduction in VBench score.

\section{Background}

\subsection{Diffusion Models}

Diffusion models~\citep{sohl2015deep, ho2020denoising, dhariwal2021diffusion} generate images by iteratively removing noise from samples drawn from a known noisy distribution. Specifically, they consist of two complementary Markov processes: a forward noising process and a reverse denoising process. In the forward process, an initial clean image $x_0$ is progressively corrupted into a purely noisy image $x_T$ across a discrete sequence of timesteps $t = 1, \dots, T$, following the Gaussian conditional distribution $q(x_t \mid x_{t-1}) = \mathcal{N}(x_t; \sqrt{1 - \beta_t}\, x_{t-1}, \beta_t I)$, where $\beta_t \in (0, 1)$ defines the noise schedule at timestep $t$.
Equivalently, this forward process can be expressed as $x_t = \sqrt{\bar{\alpha}_t}\, x_0 + \sqrt{1 - \bar{\alpha}_t}\,\epsilon$, with $\epsilon \sim \mathcal{N}(0, I)$ and $\bar{\alpha}_t = \prod_{i=1}^{t}(1 - \beta_i)$.
The reverse process, parameterized by a neural network, gradually reconstructs the original clean image from $x_T$ to $x_0$ by approximating the conditional distributions $p_\theta(x_{t-1}\mid x_t) = \mathcal{N}(x_{t-1}; \mu_\theta(x_t,t), \Sigma_\theta(x_t,t))$. 
Model training involves minimizing the discrepancy between the predicted and actual noise at each timestep, enabling accurate modeling of the underlying data distribution.

\subsection{Vision- and Diffusion Transformers} 
\label{sec:bg_vits}
Following their success in language modeling, transformers~\citep{vaswani2017attention} have been successfully applied to vision tasks. Vision transformer~\citep[ViT]{dosovitskiy2020image} achieves this by constructing a flattened token sequence from pixel patches. Similar to transformers, ViT models consist of alternating self-attention~\citep{bahdanau2014neural} and multi-layer perceptron (MLP) blocks, with intermediate layer normalization~\citep{ba2016layer}. In these models, self-attention distributes information across the sample, in this case passing information between image patches, while the MLP processes tokens independently. Diffusion transformers~\citep[DiT, ][]{peebles2023scalable} adapt ViT models for image diffusion with adaptive layer normalization (adaLN), which conditions the normalization operation on the diffusion timestep and image class.

In addition to image generation, DiTs are the predominant architecture for video generation.
Here, bidirectional video models such as Wan2.1 \citep{wan2025}, which generate entire videos at once without causal constraints, are at the forefront of generation quality.
To accelerate inference and allow for long-horizon generation, they are often finetuned into autoregressive models with fewer denoising steps 
\citep{yin2025slowbidirectionalfastautoregressive,huang2025selfforcingbridgingtraintest,zhu2026causalforcingautoregressivediffusion}.
\method{} is independent of, and compatible with these techniques.

Outside of diffusion, various extensions to ViT have been proposed. \citet{darcet2023vision} analyse the attention map of ViT models and find attention sinks: low-information image patches with anomalously high attention weights that pool global information. Therefore, the authors propose \textit{register tokens} with learned embedding values that are appended to the image sequence and discarded at the end of the forward pass.
We extend our models to use register tokens to improve compute allocation, inspired by
\citet{darcet2023vision}, who demonstrate that removing the attention sinks from the image improves the performance and alignment of saliency maps with semantic features.

\subsection{Mixture-of-Experts Models}

Mixture-of-experts~\citep[MoE,][]{shazeer2017outrageously,fedus2022switch} approaches improve transformer scaling by dividing the transformer's MLP blocks into multiple \emph{expert} MLPs, with each token typically routed to only a single expert, as determined by a routing module, a small MLP that computes scalar scores for each token-expert pair. These scores determine the assignment of tokens to experts via (potentially stochastic) top-$K$ selection.

While earlier approaches directly optimize stochastic routing decisions to minimize the expected loss using REINFORCE-style objectives \citep{bengio2015conditional}, recent work instead uses learned gating functions as the router, using the gate as a heuristic proxy for expert utility \citep{shazeer2017outrageously}.
\cite{raposo2024mixture} use such a routing mechanism for sparse token processing in language processing, and we propose combining it with dynamic MLP sparsity for image patches.

\section{Token Routing in Diffusion Transformers}
\label{sec:routing-mechanism}

\begin{figure*}[t]
    \centering
    \includegraphics[width=\linewidth]{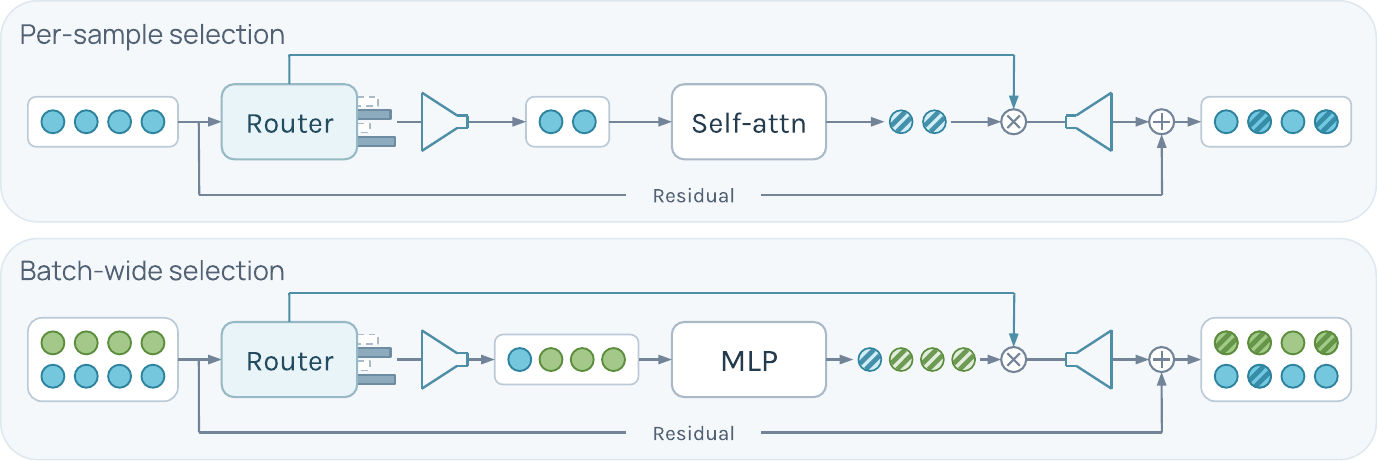}
    \caption{Token routing in \method{}. \textbf{Top:} self-attention uses fixed top-$K$ routing per sample, selecting two of four tokens at the illustrated 50\% capacity. \textbf{Bottom:} MLP routing shares the training budget across samples, selecting three green tokens and one blue token. Colors denote samples; striped overlays show additive updates. Funnels gather selected tokens and scatter their updates back to the original positions. The router gates update ($\times$) before residual addition ($+$); unselected tokens retain their original values.}
    \label{fig:arch}
\end{figure*}

\method{} combines a token bypass \citep{krause2025tread} with separate routing for attention and MLPs \citep{zhao2025dydit,zhao2026dyditpp}, allowing for dynamic allocation in MLPs without ragged batches \cite{shi2025diffmoe,yuan2025expertrace}.

\subsection{Learning to Route Tokens}
\label{ssec:routing-basics}

At the core of our method is a simple routing mechanism (see \cref{fig:arch}), similar to an MoE routing module as used in \citep{raposo2024mixture,you2025diffcr}.
Each of our router blocks consists of a tiny, two-layer MLP $R_{\phi}$  applied independently to each token $t_i$ with $i\in {1, \dots, N}$ to compute a scalar routing logit $R_{\phi}(t_i)$.
During training, we apply stochastic Gumbel top-$K$~\citep{jang2016categorical,pmlr-v97-kool19a} routing to encourage exploratory routing decisions.
Specifically, for each token we sample $g_i \sim \mathrm{Gumbel}(0,1)$ and select the $K$ tokens with the highest perturbed score,
\[
\tilde{r}_i = R_{\phi}(t_i) + \tau g_i,
\]
where $\tau$ is a temperature parameter. By annealing $\tau \to 0$, we encourage exploration during early training while recovering increasingly deterministic top-$K$ routing from the unperturbed logits.

As in MoE models, the router receives gradients by simultaneously acting as a gating mechanism.
Similarly to \citet{raposo2024mixture}, we multiply the output of the routed block for each token by its routing weight, 
\begin{equation}
    r_i = \sigma(R_{\phi}(t_i)).
\end{equation}

\subsection{Dynamic Allocation in MLP Routing}
\label{sec:mlp-routing}

MLPs process each token independently, allowing for a simple way to introduce dynamic sparsity.

\textbf{During training}, instead of selecting the top-$K$ tokens per sample, we can merge the batch and token dimensions and select the top-($B\cdot K$) tokens across the whole batch on each GPU \citep{shi2025diffmoe,yuan2025expertrace}.
Since samples in each batch belong to different classes and diffusion steps, the model can learn to dynamically allocate compute based on image class and diffusion step (\cref{sec:im-expts}).

\textbf{At inference}, however, we might want to generate a single sample ($B=1$). Selecting the top-$(1\cdot K)$ tokens would lose the dynamic allocation based on class and noise level, thereby introducing a distribution shift in the number of selected tokens.
Instead, we can approximate the global top-$(B\cdot K)$ operation applied during training by tracking a logit threshold $T$ at training time, such that $p(R_\phi(t_i) \geq T) = c$.
Here $c = K/N$ is the model capacity, i.e., the targeted fraction of tokens we route on average.
We initialize at $T = 0$ and update $T \leftarrow T + \alpha (\hat r - c)$, where $\alpha$ is a learning rate and $\hat r$ is the fraction of tokens with a clean score $R_\phi(t_i) \geq T$. 

This differs from Expert Race~\citep{yuan2025expertrace}, which averages the cutoff value to the top-$K$.

\subsection{Fixed-Capacity Routing for Self-Attention}
\label{sec:value-gating}

In self-attention, the number of tokens in an example determines the dimensions of its attention matrix.
For efficient parallel processing on GPUs, all examples in the same batch must use attention matrices with the same dimensions.
Therefore, we cannot simply give each example a different number of tokens, because this would produce differently sized matrices and prevent efficient vectorized computation.
We subsequently route a fixed number of $K$ tokens for each example.

Additionally, tokens have two distinct roles, each producing an output value and serving as context for other tokens to attend to.
This opens up a design space for routing context (keys and values) and output (queries) independently, or routing based on relevance as context.
We elaborate and show results for possible setups in \cref{sec:routing-modes}, and find that the simple method in \cref{ssec:routing-basics} is among the best, so we select it for its simplicity.

\section{Results on ImageNet}
\label{sec:im-expts}
\paragraph{Experimental Setup} We perform experiments on ImageNet~\citep{imagenet}, using the Face-blurred ILSVRC 2012–2017 variant~\citep{yang2022study} at $256\times256$ resolution. We train all models from scratch for 1M steps using the DiT-L/2 configuration and all hyperparameters from \citet{peebles2023scalable} without further tuning.
This includes using the same pre-trained variational autoencoder~\citep[VAE]{kingma2013auto} from Stable Diffusion~\citep{rombach2022high}, with an $8\times$ downsample factor.
To make \method{} comparable in FLOPs with DiT, we construct \method{} by replacing alternate DiT blocks with two \method{} blocks, approximately matching DiT FLOPs for a 50\% capacity model with the same configuration. For more details, see \cref{sec:imagenet-details}.

\paragraph{Evaluation Metrics} We evaluate the performance of image generation using standard evaluation metrics of generative models: Fr\'echet Inception Distance~\citep[FID]{heusel2017gans}, spatial FID~\citep[sFID]{nash2021generating}, Inception Score~\citep[IS]{salimans2016improved}, and Precision/Recall~\citep{kynkaanniemi2019improved}, each using 50k samples.

\begin{figure*}[t]
    \centering
    \includegraphics[width=\linewidth]{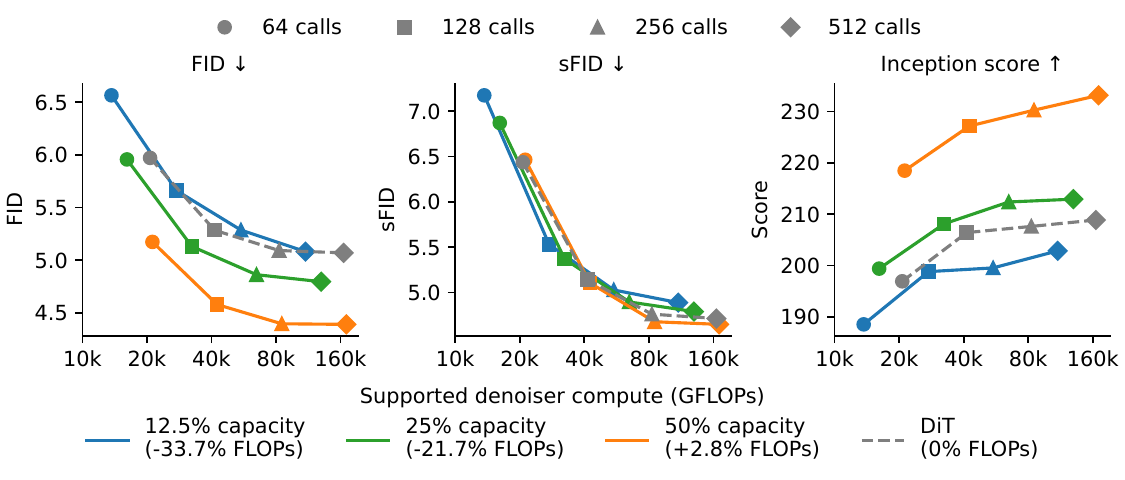}
    \caption{ImageNet performance---we report FID-50K, sFID-50K, and IS for samples generated with varying numbers of diffusion timesteps. Precision and recall evaluation is in \cref{sec:im-prec-recall}.}
    \label{fig:im-perf-flops}
\end{figure*}

\paragraph{Results} \Cref{fig:im-perf-flops} and \cref{sec:im-prec-recall} show the FLOPs-normalized performance of 50\%, 25\%, and 12.5\% \method{} models, as well as a dense DiT baseline.
The 50\% \method{} model, which we constructed to match the baseline FLOPs, consistently outperforms the DiT baseline in FID, sFID, precision, and IS. 
Notably, the 25\% model achieves better FID, sFID, and IS, at a 22\% reduction in training and inference FLOPs.
Furthermore, the 12.5\% model is competitive with DiT on all metrics, despite using 33\% less FLOPs.

\paragraph{Ablations} We experiment with multiple attention gating methods (see \cref{sec:value-gating}), and present full results in \cref{sec:routing-modes}. Our evaluations show that no method uniformly performs best. We thus settle for the simplest possible attention gating setup, as visualized in \cref{fig:arch}. Furthermore, \cref{sec:routing-ablations} shows that learned token routing substantially improves over random token selection, while independent routing does not uniformly beat coupled routing. Coupled routing has better FID, while the full method has better sFID.

\begin{figure*}[t]
    \centering
    \includegraphics[width=\linewidth]{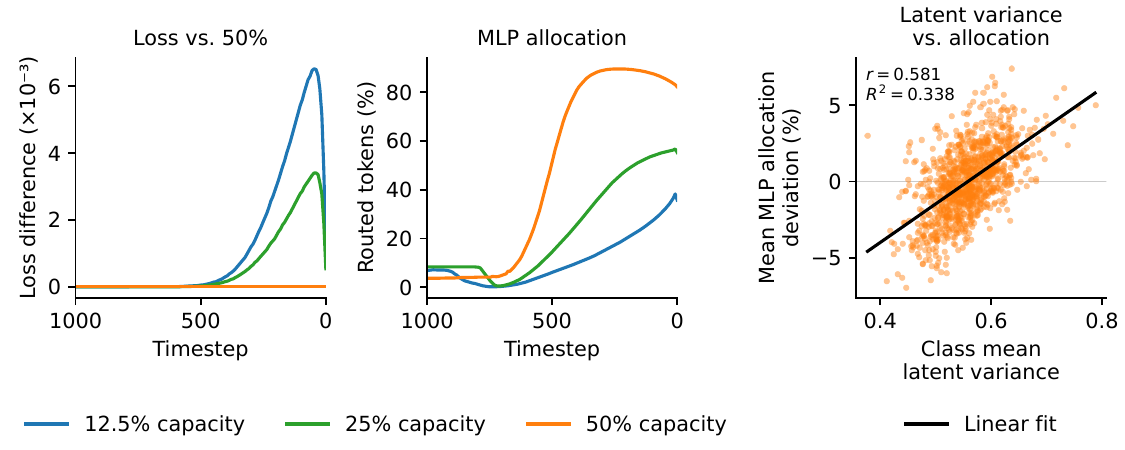}
    \caption{\method{} compute allocation increases at lower noise levels, and shows a preference towards classes with higher clean latent variance.}
    \label{fig:im-loss-capacity}
\end{figure*}

\paragraph{Compute Allocation over Noise Levels}
\cref{fig:im-loss-capacity} shows the compute allocation of samples over denoising steps, as well as the difference in diffusion loss of \method{} models with different capacity levels. Again, we show only the MLP routing proportion since the self-attention capacity is constant across samples. We see increased compute allocation at lower noise levels (smaller timesteps) for all models, as well as a greater difference in allocated capacity between models, validating the qualitative example in \cref{fig:im-sample}. The difference in loss between \method{} models is roughly proportional to the difference in compute allocation between them, suggesting models spend their increased capacity at low noise levels to decrease loss. 

\paragraph{Compute Allocation over Classes}
We analyze compute allocation of \method{} models, and find that the variance within each image's latent representation is a strong predictor of allocation (\cref{fig:im-loss-capacity}).
We hypothesize that classes with high-variance latent patches require more iterative refinement throughout the whole denoising process, for example, to update boundaries and relationships to surrounding patches. 
A uniform patch's representation contains relatively less information, so additional computation adds little bonus.
We find that lower-capacity models have weaker correlation, full results can be found in \cref{fig:class-variance-over-time}.

\section{Results on 1X}
\label{sec:vid-expts}

\paragraph{Experimental Setup} 
We use the raw video dataset from the 1X World Modeling Challenge~\citep{1x_world_model_raw_data_2025}, containing approximately 100 hours of humanoid robot POV video at 30 frames per second and $512\times512$ resolution, as well as continuous controller actions for the robot at every frame. To embed these videos for latent diffusion, we use the pre-trained Cosmos Tokenizer~\citep{agarwal2025cosmos}, selecting the continuous variant with an $8\times$ spatial and temporal downsample factor. This embeds 8 contiguous video frames with dimension $8\times 512 \times 512 \times 3$ into a latent frame with dimension $1 \times 64 \times 64 \times 16$.
During training, we condition models on two latent context frames, equivalent to 16 image frames, as well as the next 8 robot actions (corresponding to 1 latent frame). We train models to denoise the next latent frame conditioned on this context, and autoregressively sample latent frames during evaluation.

In order to support efficient training on video data, we also introduce spatio-temporal (ST) variants of DiT and \method{}, ST-DiT and ST-\method{}. These separate the self-attention operation into separate spatial and temporal attention operations, in which tokens attend to other tokens in the same frame or other tokens in the same position in \textit{previous} frames, respectively. We compare ST-DiT and ST-\method{} using the DiT-L/2 configuration at 200,000 optimizer updates (\cref{sec:1x-details}).

\paragraph{Results}
We evaluate the 200k EMA checkpoints on 400 paired validation clips, measuring PSNR at $0.5$, $1.0$ and $2.0$ seconds after the observed context (\cref{tab:vid-results}). ST-\method{} at 12.5\% capacity improves PSNR by $0.677$, $0.376$ and $0.213$\,dB, respectively; all three pointwise 95\% paired segment-bootstrap confidence intervals exclude zero. It uses 36.92\% fewer counted denoiser FLOPs on the profiled clip. Sparse and dense models use training batches of 256 and 217, respectively, so equal optimizer-update counts do not imply equal training-data exposure.

\begin{table*}[t]
    \centering
    \footnotesize
    \setlength{\tabcolsep}{4pt}
    \caption{1X video prediction at 200k EMA checkpoints on 400 paired validation clips from 249 segments. The final two rows give sparse-minus-dense deltas and their pointwise 95\% paired segment-bootstrap confidence intervals for PSNR gains. Denoiser TFLOPs cover the first clip's full rollout, including both guidance branches (\cref{sec:1x-details}). Deltas are calculated before rounding.}
    
    \begin{tabular}{@{}lrrrrr@{}}
    \toprule
        Model & Parameters & Denoiser & \multicolumn{3}{c}{PSNR (dB)$\uparrow$} \\
        & (M) & (TFLOPs)$\downarrow$ & +0.5s & +1.0s & +2.0s \\
        \midrule
        ST-DiT & 651.2 & 7,746.3 & 24.8 & 22.6 & 21.0 \\
        ST-\method{} (12.5\%) & 942.7 & 4,886.4 & {25.4} & {23.0} & {21.2} \\
        \midrule
        $\Delta$ (ST-\method{} $-$ ST-DiT) & +291.5 & $-$2,859.8 & $+$0.68 & $+$0.38 & $+$0.21 \\
        95\% CI & & & [0.486,\,0.871] & [0.229,\,0.515] & [0.075,\,0.344] \\
        \bottomrule
    \end{tabular}
    \label{tab:vid-results}
\end{table*}

\begin{figure*}[t]
    \centering
    \includegraphics[width=\linewidth]{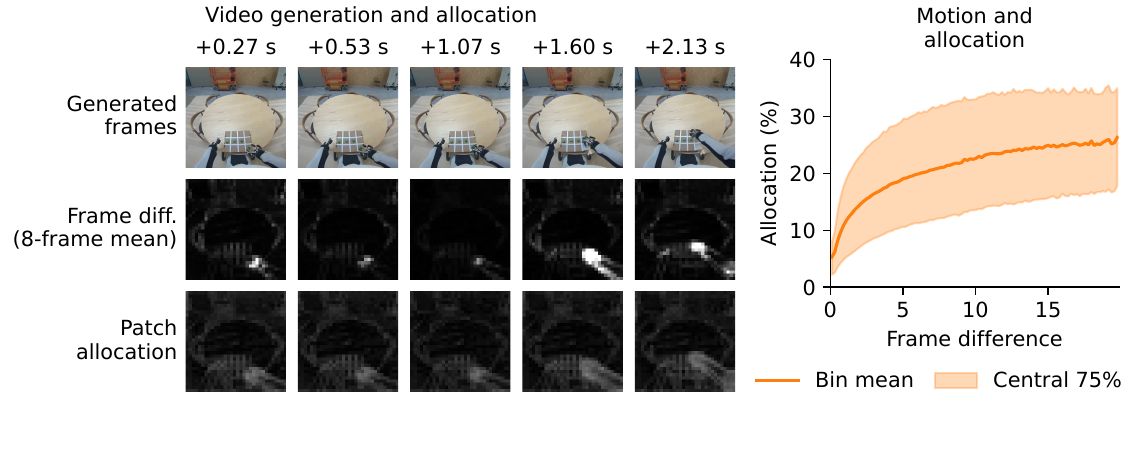}
    \caption{Compute allocation is strongly correlated with video motion---we show mean routing percentage for binned values of patch difference (clipped at the 99$^\text{th}$ percentile). Patch difference is computed as the magnitude of the difference between generated patches in adjacent frames, and the model is trained on the 1X World Modeling Dataset, with a capacity of 12.5\%.}
    \label{fig:vid-compute-alloc}
\end{figure*}

\paragraph{Compute Allocation over Temporal Pixel Difference}
Video generation models are conditioned on previous frames, making it trivial to denoise future tokens corresponding to static objects; they can just be copied over. We therefore hypothesise that dynamic objects with high optical flow should benefit from increased compute allocation in \method{} models. To evaluate this, we measure the allocation percentage and absolute pixel difference in 8 consecutive frames (corresponding to the temporal downsampling factor of the video tokenizer). We visualise the result in \cref{fig:vid-compute-alloc}, which shows a strong positive correlation between a token's frame difference and allocation frequency, confirming that \method{} uses more FLOPs to generate frames with many dynamic objects (\cref{sec:1x-motion}).

\section{Sparse Distillation with Self-Forcing}
\label{sec:distillation}
\paragraph{Experimental Setup}
We investigate the distillation of pretrained Wan2.1-T2V-14B~\citep{wan2025} into a sparse, autoregressive few-step student initialized from Wan2.1-T2V-1.3B, following Self Forcing~\citep{huang2025selfforcingbridgingtraintest}.
Our student retains the native 30-block architecture and generates 81-frame videos at $832\times480$ resolution, processing chunks of three latent frames with four denoising steps per chunk.

We apply \method{} routing to MLP sublayers and self-attention queries, retaining dense keys, values, and text cross-attention.
We found that decreasing capacity throughout the model's blocks helped preserve both global structure and fine details; see \cref{sec:sf-capacity}.
We train two models, Funnel50 and Funnel20, which respectively decrease each block's capacity from 100\% to 50\% and 20\%.
Both models use sparse attention and MLP routing.
To control stochastic exploration, we adapt the routing-noise temperature to attain a target replacement rate within the top-$K$.
The student weights and routers are jointly optimized.
We gradually introduce sparsity during adaptation and use straight-through router gates to preserve the scale of pretrained updates.
Training uses AdamW with gradient accumulation to obtain an effective batch size of 64.
More details can be found in \cref{sec:sf-details}.

\paragraph{Results}
\begin{figure*}[t]
    \centering

    \begin{minipage}[c]{0.35\linewidth}
        \centering
        \includegraphics[width=\linewidth]{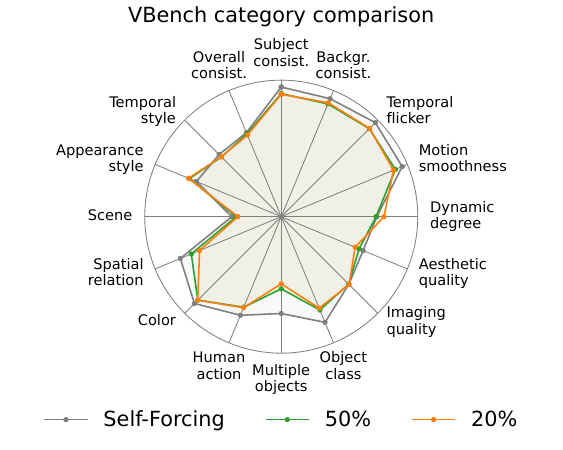}
    \end{minipage}
    \hfill
    \begin{minipage}[c]{0.62\linewidth}
        \centering
        \resizebox{0.9\linewidth}{!}{
        \setlength{\tabcolsep}{3pt}
        \begin{tabular}{lrrr}
    \toprule
     & Funnel20 & Funnel50 & Self-Forcing \\
    \midrule

    VBench Total $\uparrow$
         & 77.69 & 78.03 & 82.07 \\
    VBench Quality $\uparrow$
         & 81.07 & 81.15 & 84.89 \\
    VBench Semantic $\uparrow$
         & 64.14 & 65.54 & 70.82 \\

    Motion dynamic $\uparrow$
         & 75 & 75 & 82 \\
    Instruction following $\uparrow$
         & 98 & 100 & 98 \\

    \midrule
    
    Transformer TFLOPs $\downarrow$
         & 722.53 & 789.93 & 950.39 \\
    FLOP reduction $\uparrow$
         & 23.98\% & 16.88\% & 0\% \\

    \bottomrule
\end{tabular}
    }
    \end{minipage}

    \caption{Results of distilling a \method{} model on the VBench benchmark, as well as qualitative results compared to the dense baseline (\cref{sec:vbench-details}).}
    \label{fig:vbench}
\end{figure*}
We find that sparse \method{} models perform similarly to the dense Self Forcing baseline on the VBench benchmark, with modest reductions in most evaluation categories.
The sparse models struggle particularly in generating spatial relations and multiple objects, which require compute to be distributed across the image.
Overall, \method{} achieves a total VBench score that is $5.35\%$ lower than the dense baseline, at a $23.98\%$ reduction in transformer FLOPs (\cref{sec:vbench-details}).

\section{Related Work}
Transformers have become the predominant architecture for flow-based generative modeling, in both image generation with models such as DiT~\citep{peebles2023scalable}, DALL-E 3~\citep{goh2023dalle3}, and Latte~\citep{ma2024latte}, as well as video generation in GAIA-2~\citep{russell2025gaia2}, Genie 2~\citep{parkerholder2024genie2}, Mochi~\citep{yang2024mochi}, Sora~\citep{brooks2024sora}, MovieGen~\citep{polyak2024movie}, and Wan2.1~\citep{wan2025}. This paradigm naturally invites adaptive compute allocation, since denoising difficulty varies across noise levels and spatio-temporal regions. Furthermore, many of the downstream applications of these models, such as large-scale image generation and real-time video generation, have strict constraints on latency that require fast inference. Despite this, adaptive computation techniques have not yet been adopted in state-of-the-art models.

However, interest in this research area has recently increased. 
\citet{wang2024sparsedm} learn sparsity masks for pre-trained diffusion models without content-dependent selection, while \citet{xi2025sparsevideogen} use online profiling to sparsify spatial-temporal attention connections while retaining all tokens.
Token compression methods reduce the sequence length through token merging in ToMeSD~\citep{bolya2023tokenmergingfaststable}.
ToCa~\citep{zou2025acceleratingdiffusiontransformerstokenwise} and RAS~\citep{liu2026regionadaptivesamplingdiffusiontransformers} reuse computation across denoising timesteps through cached token features and noise predictions for inactive regions, respectively.
DiffCR~\citep{you2025diffcr} routes tokens through or around whole transformer blocks and learns layer- and timestep-specific capacities with an auxiliary budget loss, evaluated by fine-tuning pretrained diffusion models.

More closely related, DyDiT~\citep{zhao2025dydit} and DyDiT++~\citep{zhao2026dyditpp} combine timestep-conditioned width reduction with content-dependent token bypass in MLP blocks, using an auxiliary loss to control overall FLOPs.
They require full self-attention during training, so sparsity is only introduced in inference.
TREAD~\citep{krause2025tread} accelerates training by randomly selecting tokens to bypass sequences of transformer blocks, while retaining dense computation at inference.
Our method learns separate token routers for attention and MLP blocks jointly with the diffusion model from scratch, without auxiliary losses, while preserving tensor shapes for efficient batch inference with per-sample flexibility in the MLP blocks.

Mixture-of-experts mechanisms~\citep[MoE]{shazeer2017outrageously,fedus2022switch} allocate computation by routing tokens among multiple MLP experts, as in EC-DiT~\citep{sun2025ecdit}.
DiffMoE~\citep{shi2025diffmoe} pools tokens across examples and learns a capacity predictor for adaptive inference; Expert Race~\citep{yuan2025expertrace} combines global token–expert competition with tracked-threshold inference. 
MoDDiT shares these allocation principles but applies them to a single MLP branch, combines them with independently routed sparse attention, and trains the model from scratch without a separately supervised capacity predictor or auxiliary routing objectives.

Outside of diffusion modeling, adaptive computation has become increasingly prevalent in transformer models, in order to allocate resources efficiently based on input complexity.
\citet{xin2020deebert} propose a language model architecture with adaptive depth via an early exiting mechanism.
Inspiring this work, mixture-of-depths~\citep{raposo2024mixture} uses a routing mechanism to limit the capacity of transformer layers in autoregressive language models. While this method shows improvements in the FLOPs-normalized performance frontier of such models, we find diffusion to be an even more suitable paradigm for this architecture, as the models avoid causal structure interfering with the top-$K$ operation.

\section{Conclusion}

\paragraph{Summary}
\method{} represents a significant advancement in efficient generative modeling, effectively addressing a computational inefficiency inherent in existing diffusion models by dynamically allocating compute across tokens, samples, and noise levels. By employing a minimal yet powerful routing mechanism, \method{} demonstrates considerable performance improvements at the same level of compute, on both image and video generation tasks. Our evaluation also provides an intuitive understanding of the learned routing mechanism, highlighting \method{}'s capability to align computational effort with semantic importance and temporal dynamics. Our work thus provides a step towards unlocking generative modeling in latency- and hardware-constrained settings and offers insights into the distribution of computational resources within diffusion models.

\paragraph{Limitations and Future Work}

While our work provides an extensive evaluation, it is ultimately limited by our computational resources, providing the opportunity for further evaluation with alternate datasets, architectures, and generative algorithms.
Namely, image generation datasets such as ArtBench~\citep{liao2022artbench} and Food~\citep{bossard2014food}, as well as video generation datasets such as Taichi-HD~\citep{siarohin2019first} and Minecraft VPT~\citep{baker2022video}, would strengthen confidence in the performance of \method{}.

While \method{} can substantially reduce training and inference FLOPs, this reduction does not automatically translate into wall clock time savings.
Even though we observe throughput gains at larger batch sizes \footnote{At batch size 32, we measure 1.64 images/sec for a 50\% \method{} model, vs. 1.55 images/sec for a DiT.}, generating few images in parallel is slower.
This is likely due to a combination of the additional overhead of the gather and scatter operations, and the fact that dense matrix multiplication is already highly optimized itself.
Future work can explore the development of optimized kernels for sparse \method{} blocks, which have proven to lead to significant performance increases for attention blocks \citep{dao2023flashattention2}.

Finally, \method{} introduces a capacity hyperparameter for the MLP and self-attention blocks, providing a range of promising avenues for further investigation. While our method avoids auxiliary losses, the addition of such a loss could enable automatic tuning of this capacity, with a regularization penalty against high compute budgets. Furthermore, we use the same capacity for both MLP and self-attention blocks. Visual inspection suggests that they have distinct allocation dynamics (\cref{fig:im-sample}), meaning their optimal capacities likely differ and should be investigated.

\newpage
\section*{Contributions}
JL and MJ co-led the project. MJ developed and implemented the initial method, ran experiments on ImageNet and 1X, and wrote the \href{https://github.com/EmptyJackson/personal-site-two-col/blob/bc537f8c442aa154536564c081be7e1ef9ffd789/static/sparse_dit.pdf}{first version of the paper} (May 2025). JL ran extensive ablations, modified and extended the method, added experiments including sparse distillation, and polished the manuscript. JF, SW, and BE co-supervised the project.

\bibliography{refs}
\bibliographystyle{icml2026}

\clearpage
\appendix
\onecolumn
\crefalias{subsection}{appendix}
\crefalias{subsubsection}{appendix}
\section{ImageNet experiments}

\subsection{Training and evaluation}
\label{sec:imagenet-details}

\paragraph{Dataset} We perform experiments on ImageNet~\citep{imagenet}, using the Face-blurred ILSVRC 2012–2017 variant~\citep{yang2022study} at $256\times256$ resolution.

\paragraph{Model Architecture} We train latent diffusion models using the pre-trained variational autoencoder~\citep[VAE]{kingma2013auto} from Stable Diffusion~\citep{rombach2022high}, with an $8\times$ downsample factor, giving a latent embedding with dimensions $32\times 32 \times 4$.
Our baseline is the DiT-L/2 architecture from \citet{peebles2023scalable}. 
To make \method{} comparable in FLOPs with DiT, we construct \method{} by replacing alternate DiT blocks with two \method{} blocks, approximately matching DiT FLOPs for a 50\% capacity model with the same configuration.
Finally, \method{} models append every latent image with 16 register tokens \citep{darcet2023vision}, as we found this to help with aligning allocation masks with semantic features, giving a total embedding length of 272.
We initialize the final router layer at zero, such that the output is constant $0.5$.
We use a cosine decay for router Gumbel noise, decaying from $5$ to $0$ over training.

\paragraph{Training} We train all models from scratch for 1M steps (requiring approximately 6 days on $4$ L40S GPUs) using hyperparameters from \citet{peebles2023scalable} without further tuning. 
This includes using a global batch size of 256 with AdamW using a learning rate of $10^{-4}$, $\beta = (0.9, 0.999)$, and no weight decay.
We use an EMA value of $0.9999$ throughout training.

\paragraph{Evaluation} We evaluate all models using $50,000$ generated images at $64, 128, 256,$ and $512$ denoising steps and a classifier-free guidance scale of $1.5$. We measure FLOPs on the first 1024 generated images, including both classifier-free guidance branches, excluding the VAE decoder. 

\subsection{Additional results}
\label{sec:im-prec-recall}

\begin{figure}[htbp]
    \centering
    \includegraphics[width=\linewidth]{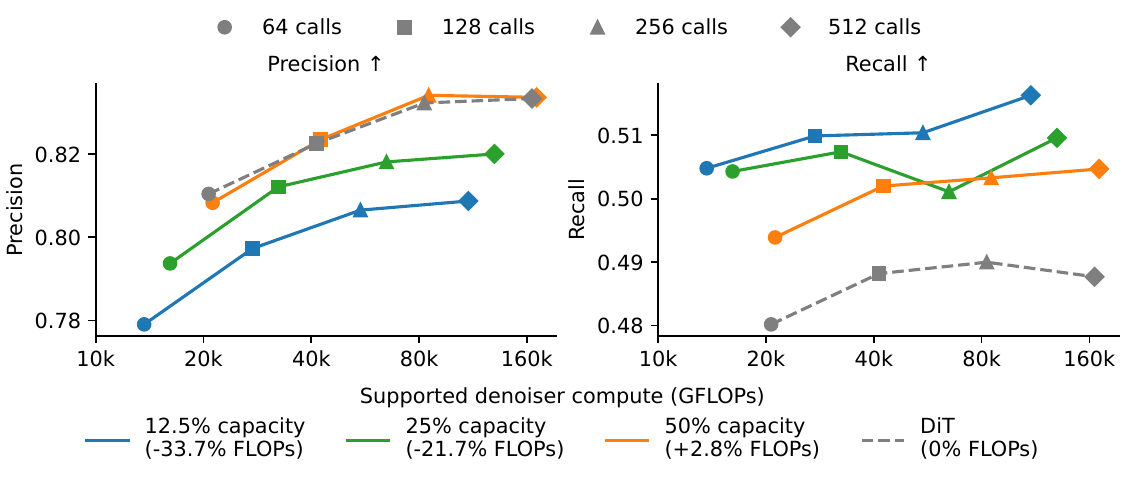}
    \caption{ImageNet performance.}
    \label{fig:im-prec-recall}
\end{figure}

\begin{table}[htbp]
\centering
\caption{Evaluation at 512 DDPM steps with CFG scale 1.5.
Quality metrics use 50,000 images; compute estimates average
1,024 images and includes both CFG branches.}
\label{tab:imagenet-flops}
\begin{tabular}{lrrrrrr}
\hline
Model & TFLOPs/image & FID $\downarrow$ & sFID $\downarrow$
& IS $\uparrow$ & Precision $\uparrow$ & Recall $\uparrow$ \\
\hline
DiT-L/2           & 165.380 & 5.069 & 4.714 & 208.861 & \textbf{0.834} & 0.488 \\
\method{} (50\%)   & 170.316 & \textbf{4.390} & \textbf{4.649} & \textbf{233.120} & \textbf{0.834} & 0.505 \\
\method{} (25\%)   & 129.888 & 4.795 & 4.790 & 212.914 & 0.820 & 0.510 \\
\method{} (12.5\%) & 109.783 & 5.081 & 4.889 & 202.856 & 0.809 & \textbf{0.516} \\
\hline
\end{tabular}
\end{table}

\begin{figure}[htbp]
    \centering
    \includegraphics[width=\linewidth]{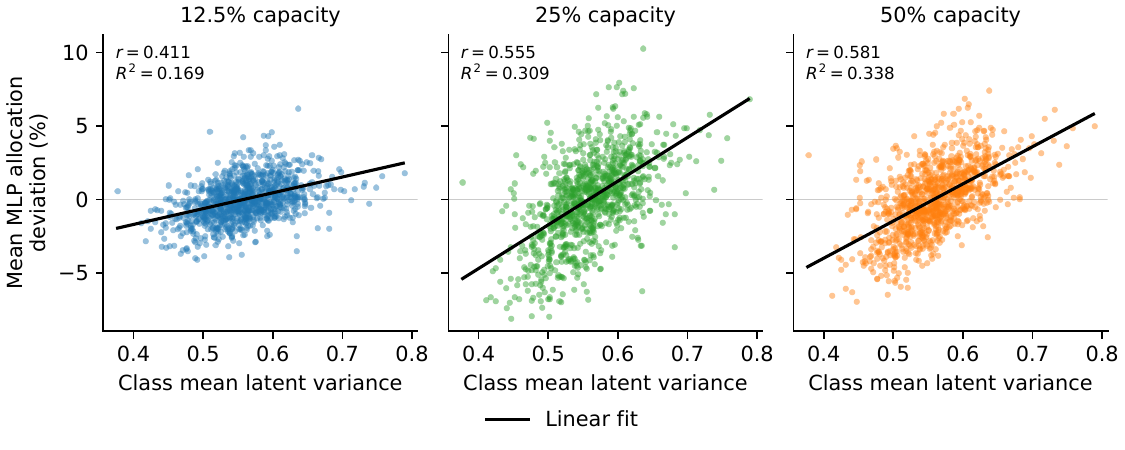}
    \caption{Allocation is consistently correlated with mean latent variance of each class, and the correlation becomes smaller for smaller capacities.}
    \label{fig:class-variance-over-time}
\end{figure}

\subsection{Routing design studies}

\subsubsection{Attention-routing alternatives}
\label{sec:routing-modes}

Gating values serve as a proxy for token selection.
But what form of gating produces the most useful proxy?
Multiplying the attention output by the router score incentivises retaining tokens whose outputs reduce loss. 
However, there is no incentive to increase the scores of tokens that serve as useful context for others to attend to.
To provide such a signal, we may additionally gate the attention values using the routing scores.
Let $A = \operatorname{softmax}(QK^T / \sqrt d)$ and $R = \operatorname{diag}(r)$. While standard output gating computes $R \operatorname{attn}(x) = RAV$, additionally gating the attention value yields
$$
R A R V = (rr^\top \odot A) V \quad \text{where} \quad (rr^\top \odot A)_{ij} = r_i a_{ij} r_j.
$$

This formulation introduces complementary incentives to increase the score $R_\phi(t_i)$ of a token $t_i$:
\begin{itemize}
\item \textbf{Output gating:} Does $t_i$ reduce downstream loss by attending to other tokens?
\item \textbf{Value gating:} Do other tokens decrease downstream loss by attending to $t_i$?
\end{itemize}

Disentangling both incentives by producing two routing scores opens a substantial design space. For example, consolidating both scores for top-$K$ selection can be done as a sum, product, or harmonic mean. 
Additionally, instead of consolidating scores, one may select two different top-$K$ subsets of tokens for $Q$ and $K$/$V$, and compute cross-attention, or implement sparsity only in keys and values. 

We test four of these configurations:
\begin{itemize}
\item \textbf{Output-only}: the default \method{}, gating only the self-attention output via $RAV$.
\item \textbf{Shared value/output}: gating the output and value by the same routing score via $RARV$.
\item \textbf{Separate routers}: two routers per attention operation, producing $R_O$ and $R_V$, and selecting tokens independently, then gating as $R_OAR_VV$.
\item \textbf{Sparse Q/dense KV}: All tokens enter attention as context, but only selected tokens contribute to the output.
\end{itemize}

\begin{figure}[htbp]
\centering
\includegraphics[width=\linewidth]{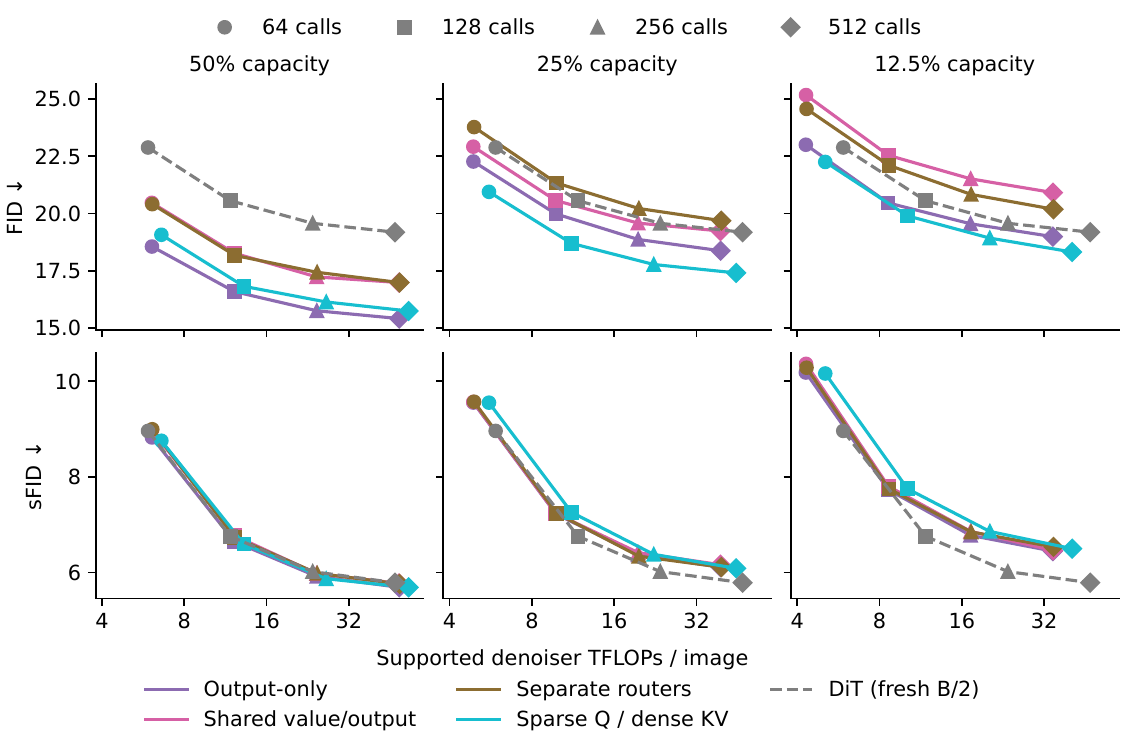}
\caption{Earlier routing sweep: FID and sFID versus counted denoiser arithmetic across sampling lengths and nominal capacities, using 10,000 images per point at 64, 128, 256, 512 denoising steps with classifier-free guidance scale of $4.0$.}
\label{fig:attention-routing-compute}
\end{figure}

\subsubsection{Selection and capacity ablations}
\label{sec:routing-ablations}

\begin{figure}[htbp]
    \centering
    \includegraphics[width=\linewidth]{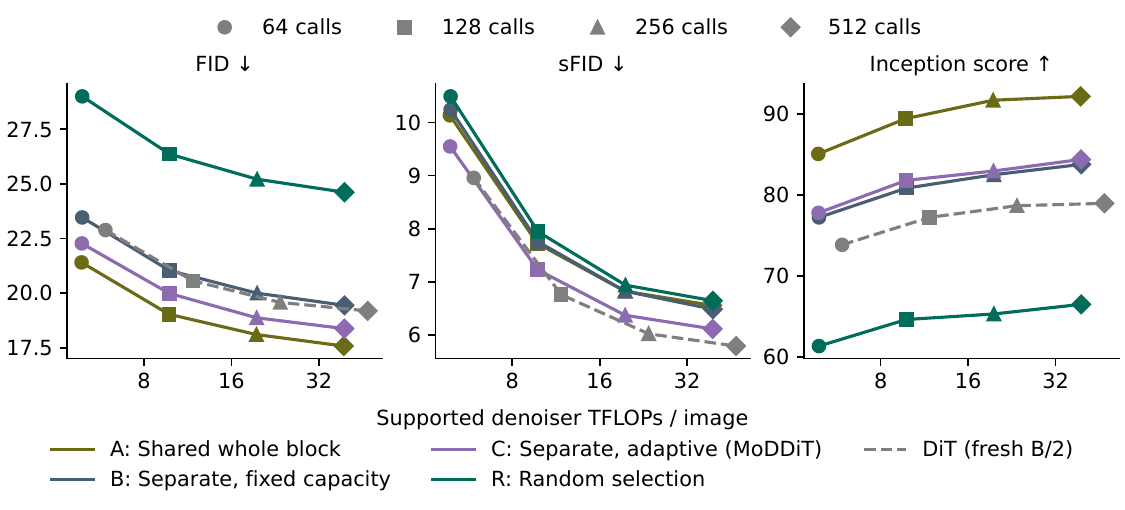}
    \caption{Routing selection and capacity ablations. Shared whole-block routing, separate fixed-capacity routing, separate adaptive routing and random selection use 25\% \method{} models trained for 400k updates from a common initialization.}
    \label{fig:routing-selection-capacity}
\end{figure}

\clearpage
\section{1X video prediction}

\subsection{Training and evaluation}
\label{sec:1x-details}

\paragraph{Dataset} We use the 1X World Modeling Dataset~\citep{1x_world_model_raw_data_2025}, with $512\times512$ video at 30\,Hz and 25-dimensional robot states.

\paragraph{Model architecture} We use the $12.5\%$-capacity DiT-L/2, block-replacement scheme and register count from \cref{sec:imagenet-details}. The frozen Cosmos CV8x8x8 tokenizer~\citep{agarwal2025cosmos} produces $64\times64\times16$ latent frames. Each frame contains 1024 spatial tokens and eight state tokens in addition to the registers. We adopt spatiotemporal attention from \cite{russell2025gaia2}, in which a token at a fixed spatial position and timestep attends to all tokens at the same timestep, and all tokens at the same spatial position in the past.

\paragraph{Training} We train from scratch with seed 0, using the AdamW settings and weight EMA from \cref{sec:imagenet-details}. Global batch sizes are 256 for ST-\method{} and 217 for ST-DiT. The learning rate warms up for 5,000 steps, then follows a cosine schedule towards $10^{-5}$ at 1M steps. Routing uses Gumbel-sigmoid temperature 5.

\paragraph{Inference} We encode the first 17 frames, then autoregressively predict eight latents (64 frames), retaining three context latents and conditioning on the next eight recorded robot states. Sampling uses 128 DDPM steps per latent and the classifier-free guidance scale from \cref{sec:imagenet-details}.

\paragraph{Evaluation} For \cref{tab:vid-results}, we evaluate both models' EMA checkpoints at exactly 200,000 optimizer updates on 400 paired, non-overlapping 81-frame windows from 249 segments in one video from the raw validation v2.0 split. Clip identities and sampling seeds are shared across models. PSNR is computed against raw RGB frames at 0.5, 1 and 2 seconds after the observed context, averaging squared error across pixels and channels before conversion to dB. We report the clip-weighted mean and paired sparse-minus-dense gains. Pointwise 95\% percentile confidence intervals use 20,000 paired bootstrap draws over segments, with seed 20261011, retaining all windows within each sampled segment. These intervals reflect evaluation variability, not training-seed variability. The temporal-change analysis in \cref{sec:1x-motion} uses its separately saved 300k, 100-clip cohort.

We count denoiser FLOPs over the first clip's complete eight-latent rollout (128 DDPM calls per latent, CFG 1.5), including both guidance branches and all dense and routed blocks. Registered matrix, convolution and attention operations count a multiply-add as two FLOPs; tokenizer, scalar/nonlinear, normalization, selection and I/O work are excluded. The table reports total rollout TFLOPs, not per-call GFLOPs or measured throughput.

\subsection{Temporal change and allocation}
\label{sec:1x-motion}

Using the setup in \cref{sec:1x-details}, we measure temporal change by averaging absolute adjacent-frame grayscale differences over each latent's temporal interval and spatial token footprint, including interval boundaries. For each spatial patch, allocation is the mean of its binary selection mask over attention and MLP branches, routed blocks, and denoising steps, expressed as a percentage. We use only the conditional guidance branch and exclude state/register tokens and dense blocks.

For \cref{fig:vid-compute-alloc}, we clip frame differences at the global 99th percentile and group them into 100 equal-width bins. The line shows mean allocation; shading spans the within-bin 12.5th--87.5th percentiles, describing observed variation rather than uncertainty in the mean. Higher temporal change is associated with greater allocation, but frame difference is not optical flow, and this association does not establish a causal benefit from routing.

\clearpage
\section{Self-Forcing adaptation}

\subsection{Sparse architecture and adaptation}
\label{sec:sf-details}
\label{sec:sf-capacity}

We follow the chunk-wise Self-Forcing DMD protocol~\citep{huang2025selfforcingbridgingtraintest}, including its prompt data, optimization settings, and inference procedure, with the following sparse adaptation details.

\paragraph{Model architecture} Starting from the released ODE checkpoint, we keep every third block dense and add separate routers of hidden width 256 for self-attention queries and MLP tokens in the remaining blocks. Keys, values, and text cross-attention remain dense, and we use straight-through gates to keep the output magnitude of the pretrained model. Attention uses fixed-count selection; MLP selection is fixed-count for Funnel20 and threshold-based for Funnel50 (\cref{sec:sf-capacity}).

\paragraph{Capacity schedules}
We experiment with two \method{} models with capacity schedules we call \textbf{Funnel20} and \textbf{Funnel50}. Across the 30 blocks, target capacity decreases linearly from 100\% at block 1 to 20\% or 50\% at block 30, respectively. Blocks 1, 4, 7, \ldots, 28 override this schedule and remain fully dense. The names are chosen as the final-block targets, not the average model-wide capacity.

Initial qualitative experiments showed that increasing capacity in early blocks could recover structure that was otherwise lost under aggressive sparsity. We also retain dense keys and values, including cached context, so selected queries can attend to all available tokens, including tokens that receive no attention update.

\paragraph{Training} We jointly adapt the student and routers for 3,000 ODE updates, reducing capacity to the target with a cosine schedule over the first 600 updates. Trajectories use 48 sampling steps and guidance scale 6, retaining five states each. We then run 25 critic warmup iterations and 1,000 DMD iterations, starting student EMA at iteration 200.

\paragraph{Evaluation} We compare the final student EMA with the released dense EMA checkpoint from Self Forcing. VBench and compute protocols are in \cref{sec:vbench-details}; qualitative comparisons are in \cref{sec:sf-qualitative}.

\subsection{VBench evaluation and compute accounting}
\label{sec:vbench-details}

\paragraph{Evaluation} We evaluate the released dense Self-Forcing EMA checkpoint and the Funnel50/Funnel20 EMA checkpoints after 1,000 DMD iterations using the sampling settings in \cref{sec:sf-details}. We use the original English VBench prompts~\citep{huang2023vbenchcomprehensivebenchmarksuite}, with five samples per prompt and reproducible seeds derived from the prompt and sample index. We apply the official category normalization and weights, and the total score weights quality and semantic scores 4:1. The separate motion and instruction scores in \cref{fig:vbench} use 100 Causal-Forcing motion prompts, one sample each: motion is the percentage classified as dynamic, and instruction following is 100 times the mean VisionReward yes/no judgment, encoded as 1 or $-1$.

\paragraph{Scores} \Cref{tab:vbench-details} gives the category scores and aggregates. Relative to dense Self-Forcing, Funnel50 and Funnel20 reduce the total score by 4.93\% and 5.35\%, respectively, while decreasing inference FLOPs by 16.8\% and 24.0\%. Both lose more on multiple objects and spatial relationships than on imaging quality. This could be because they concentrate compute on a single part of the image, not capturing relationships between different image parts.

\begin{table}[htbp]
\centering
\small
\caption{VBench scores for dense Self-Forcing and percentage-point changes for sparse EMA models relative to it (positive is better). Category scores are raw values multiplied by 100; aggregates use official normalization and weighting. Deltas are computed before rounding; categories are ordered by the Funnel20 point reduction, largest first. The final row gives absolute transformer TFLOPs for the matched clip described below (lower is better).}
\label{tab:vbench-details}
\begin{tabular}{l|r|rr}
\toprule
Metric & Self-Forcing & Funnel50 & Funnel20 \\
\midrule
Total & 82.07 & $-4.04$ & $-4.39$ \\
Quality & 84.89 & $-3.73$ & $-3.82$ \\
Semantic & 70.82 & $-5.29$ & $-6.68$ \\
\midrule
Multiple objects & 70.93 & $-18.05$ & $-21.71$ \\
Spatial relationship & 80.03 & $-8.72$ & $-15.26$ \\
Object class & 83.80 & $-9.79$ & $-11.27$ \\
Aesthetic quality & 64.92 & $-3.22$ & $-6.35$ \\
Human action & 78.20 & $-6.40$ & $-6.00$ \\
Subject consistency & 95.70 & $-4.11$ & $-4.52$ \\
Scene & 29.78 & $-2.65$ & $-3.68$ \\
Color & 89.94 & $-3.49$ & $-3.43$ \\
Background consistency & 95.22 & $-3.06$ & $-2.42$ \\
Temporal flickering & 99.04 & $-2.32$ & $-2.20$ \\
Motion smoothness & 98.56 & $-1.58$ & $-2.01$ \\
Temporal style & 23.44 & $-0.90$ & $-0.85$ \\
Overall consistency & 24.26 & $-0.36$ & $-0.71$ \\
Imaging quality & 70.02 & $-0.14$ & $+0.32$ \\
Appearance style & 19.23 & $+1.59$ & $+1.78$ \\
Dynamic degree & 70.00 & $-0.56$ & $+5.28$ \\
\midrule
Transformer TFLOPs $\downarrow$ & 950.39 & 789.93 & 722.53 \\
\bottomrule
\end{tabular}
\end{table}

\paragraph{Compute} We count matrix-operation FLOPs for one matched 81-frame raccoon-guitar clip at $832\times480$, seed 0 and batch size 1, including 28 denoising calls and six KV-cache updates. A multiply-add counts as two FLOPs. Dense Self-Forcing and Funnel20 use runtime operator shapes; Funnel50 uses recorded query and realized MLP token counts. Transformer costs are 950.39, 789.93 and 722.53 TFLOPs for dense, Funnel50 and Funnel20, giving reductions of 16.88\% and 23.98\%. Adding the shared text-encoder and VAE matrix/convolution cost of 289.39 TFLOPs reduces these savings to 12.94\% and 18.38\%. Counts exclude scalar/nonlinear operations, indexing, memory transfers and hardware padding. They describe arithmetic for this clip, not benchmark-wide averages or measured speedups; dynamic MLP allocation varies with the input.

\subsection{Qualitative comparisons}
\label{sec:sf-qualitative}

\begin{figure}[htbp]
    \centering
    \includegraphics[width=\linewidth]{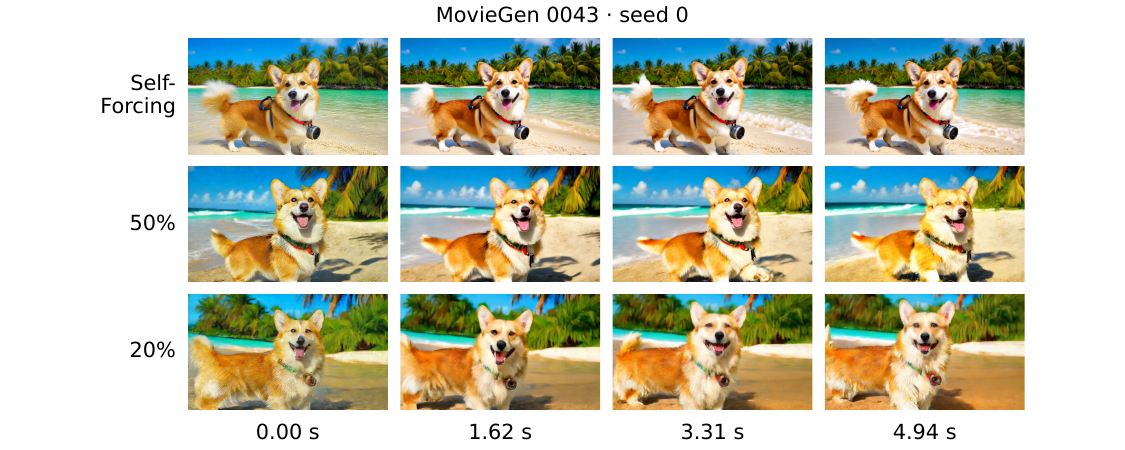}
    \caption{Matched MovieGen prompt 0043, seed 0, for dense Self-Forcing, Funnel50 and Funnel20. Columns show frames 0, 26, 53 and 79 at 16 fps.}
    \label{fig:wan-capacity-frames}
\end{figure}

\end{document}